\documentclass[journal]{IEEEtran}
\usepackage{amsmath}   % for equations
\usepackage{amssymb}   % for math symbols
\usepackage{braket}    % for \ket{} and \bra{}
\usepackage{graphicx}
\usepackage{algorithm}      % floating environment
\usepackage{algorithmicx}   % core package
\usepackage{algpseudocode}  % pseudocode syntax (for \State, \For, etc.)

\begin{document}

\pagestyle{empty}
\thispagestyle{empty}

\title{Quantum-Enhanced Generative Models for Rare Event Prediction}

\author{M.Z. Haider$^{1}$, M.U. Ghouri$^{2}$, Tayyaba Noreen$^{1}$, M. Salman$^{3}$ \\
$^{1}$Department of Software Engineering(ÉTS), Université du Québec, Canada \\
$^{2}$Department Of Computational Sciences,The University of Faisalabad, Pakistan
\\
$^{3}$Department of Computer Science, SZABIST University
}

\maketitle

\thispagestyle{empty}

\begin{abstract}
Rare events such as financial crashes, climate extremes, and biological anomalies are notoriously difficult to model due to their scarcity and heavy-tailed distributions. Classical deep generative models often struggle to capture these rare occurrences, either collapsing low-probability modes or producing poorly calibrated uncertainty estimates. In this work, we propose the \textbf{Quantum-Enhanced Generative Model (QEGM)}, a hybrid classical quantum framework that integrates deep latent-variable models with variational quantum circuits. The framework introduces two key innovations: (i) a hybrid loss function that jointly optimizes reconstruction fidelity and tail-aware likelihood, and (ii) quantum randomness–driven noise injection to enhance sample diversity and mitigate mode collapse. Training proceeds via a hybrid loop where classical parameters are updated through backpropagation while quantum parameters are optimized using parameter-shift gradients. We evaluate QEGM on synthetic Gaussian mixtures and real-world datasets spanning finance, climate, and protein structure. Results demonstrate that QEGM consistently reduces tail KL-divergence by up to 50\% compared to state-of-the-art baselines (GAN, VAE, Diffusion), while improving rare-event recall and coverage calibration. These findings highlight the potential as a principled approach for rare-event prediction, offering robustness beyond what is achievable with purely classical methods.
\end{abstract}


\IEEEpeerreviewmaketitle

%======================================================================
\section{Introduction}
%======================================================================

Rare events, including financial crashes, extreme climate phenomena, cybersecurity breaches, and rare biological anomalies, exert a disproportionate impact relative to their frequency. Accurately modeling such events is essential for effective risk assessment and informed decision-making across diverse domains\cite{lamichhane2025quantum}. For instance, systemic financial stability depends on anticipating sudden market downturns~\cite{gudivada2021quantitative}, climate adaptation strategies rely on robust models of extreme environmental changes~\cite{raj2021rare}, intrusion detection systems must recognize uncommon but critical patterns of malicious activity~\cite{zhang2022deep}, and medical research on rare diseases benefits from accurately characterizing sparse biological anomalies~\cite{li2023deep}. Traditional statistical approaches, however, often underestimate the probability of tail events, which can result in catastrophic mispredictions. Recent advances in generative artificial intelligence provide new possibilities for simulating low-probability scenarios, thereby strengthening resilience and planning in these sensitive domains\cite{tomar2025comprehensive}. Despite their success in synthesizing high-dimensional data across fields such as vision, language, and bioinformatics~\cite{dhariwal2021diffusion, razavi2022generative}, classical generative models including Generative Adversarial Networks (GANs), Variational Autoencoders (VAEs), and diffusion-based architectures face inherent limitations when modeling rare events. These methods remain biased toward frequent patterns in the training data, resulting in the poor capture of rare event distributions\cite{rattan2025quantum}. GANs frequently experience mode collapse, eliminating infrequent yet critical modes from generated samples. VAEs suffer from posterior collapse in sparse data regions, while diffusion models require significant computational resources to adapt effectively to extreme outliers~\cite{yu2023comprehensive}.

% was: \begin{figure}[h]
\begin{figure}[!t]
  \centering
  \includegraphics[width=0.95\linewidth]{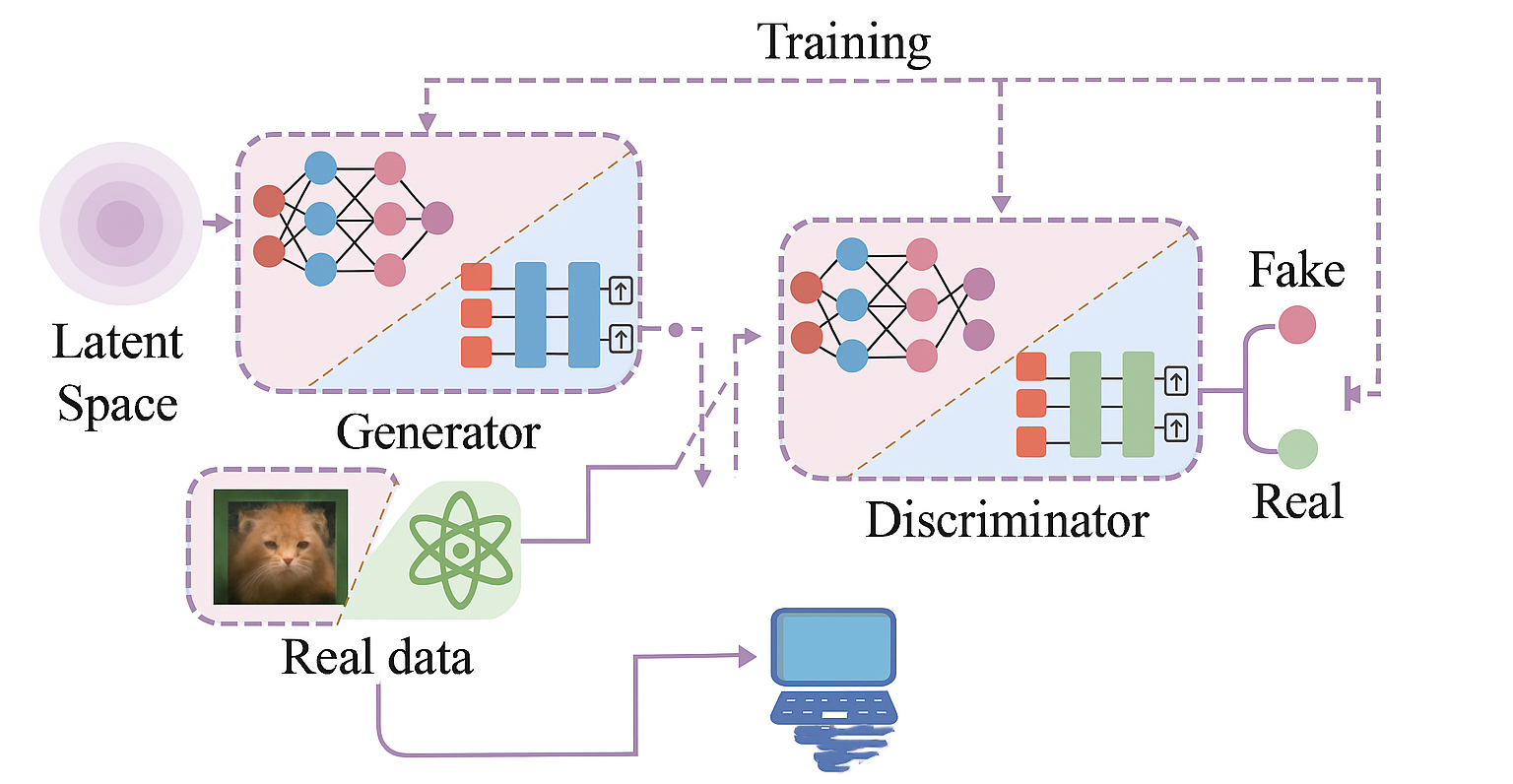}
  \caption{Real and fake data from generative models\cite{belis2024quantum}}
  \label{fig:baseline_comparison}
\end{figure}

Quantum computing introduces fundamentally different representational and sampling mechanisms that may address these shortcomings \cite{naik2025portfolio}. By exploiting superposition, quantum systems can encode exponentially many states simultaneously, and through entanglement, they can model correlations that are often difficult for classical networks to capture\cite{haider2025v}. These unique features create opportunities to advance generative modeling, particularly in sampling from rare or tail distributions that remain underrepresented in classical systems~\cite{schuld2021machine, benedetti2021variational}. Hybrid quantum classical generative models, in particular, offer the promise of overcoming classical biases by embedding quantum circuits explicitly designed to explore low-probability states. Our research makes the following contributions:
\begin{enumerate}
    \item We formalize rare event prediction as a generative modeling problem and introduce tail sensitive loss functions to better capture extreme distributions.  
    \item We design a protocol that embeds variational quantum circuits into diffusion sampling to improve coverage of low-probability states.  
    \item We provide empirical evidence on synthetic and real-world datasets,  showing QEGM achieves higher tail recall and distributional coverage than GAN, VAE, and diffusion baselines.  
\end{enumerate}
The remaining sections are organized as follows. Section~\ref{sec: background} presents the background and related work. Section~\ref{sec: problem} defines the problem formulation. Section~\ref{sec: protocol} introduces our proposed protocol, followed by Section~\ref{sec: methodology}, which describes the methodology. Section~\ref{sec: evaluation} reports the evaluation results, and Section~\ref{sec: conclusion} concludes the paper.  

%======================================================================
\section{Background and Related Work}
\label{sec: background}
%======================================================================

Rare events are low-probability but high-impact occurrences, such as financial crashes, extreme weather phenomena, and rare system failures\cite{belis2024quantum}. Unlike common patterns, these events lie in the tails of probability distributions, which makes them inherently difficult to detect and predict. Traditional statistical methods, including extreme value theory (EVT), often fail to generalize when event frequencies are sparse~\cite{faranda2021predicting}. Furthermore, imbalanced datasets bias machine learning models toward frequent cases, leading to poor recall for rare but critical events~\cite{chandola2022anomaly}. As a result, the reliable prediction of rare events remains an open challenge in artificial intelligence and risk modeling. Deep generative models have been investigated to mitigate data imbalance by synthesizing rare-event samples~\cite{10113742}. Figure~1 illustrates this advancement, showing how QGAN generates synthetic data alongside real data. Generative Adversarial Networks (GANs) have gained traction for anomaly detection and rare event augmentation, but they frequently suffer from mode collapse, where infrequent patterns vanish from the learned distribution~\cite{zhang2021ganomaly}. Variational Autoencoders (VAEs) provide probabilistic latent representations but tend to experience posterior collapse in sparse regions, which limits their effectiveness in capturing rare tails~\cite{sohn2023variational}. Diffusion models, which iteratively denoise latent variables, outperform GANs and VAEs in terms of sample fidelity, but they remain computationally intensive and biased toward dominant modes in the training data~\cite{kong2023diffusion}. These limitations highlight the need for enhanced or hybrid methods capable of representing extreme distributional behaviors.  

Quantum machine learning (QML) offers a promising alternative by exploiting superposition and entanglement to represent high-dimensional distributions more efficiently. Variational Quantum Circuits (VQCs) enable parameterized encoding of probability amplitudes, making them suitable for expressive generative modeling~\cite{schuld2021machine}. Quantum GANs (QGANs) have been introduced as quantum analogues of classical GANs, showing potential benefits in distribution sampling and optimization landscapes~\cite{huang2021experimental}. More recently, quantum diffusion models have emerged as a frontier, incorporating quantum noise injection to diversify sample generation~\cite{bao2022quantum}. Although such methods are currently constrained to small-scale datasets due to the limitations of Noisy Intermediate-Scale Quantum (NISQ) devices, they demonstrate the potential for quantum-enhanced generative modeling\cite{hibat2024framework}.  

Despite these advancements, important gaps remain in current research. Classical models continue to struggle with accurately capturing the rare tails of distributions, particularly in highly imbalanced settings\cite{haider2025range}. Quantum generative models such as QGANs have so far been validated primarily on toy or low-dimensional datasets, leaving their application to real-world rare event prediction largely unexplored \cite{rudolph2024trainability}. Moreover, the integration of quantum circuits with state-of-the-art generative pipelines like diffusion models is still underdeveloped. These gaps motivate our proposed \emph{Quantum-Enhanced Generative Model (QEGM)}, which combines quantum sampling with diffusion-based refinement to more effectively capture rare but critical events\cite{nokhwal2024quantum}.

%======================================================================
\section{Problem Formulation}
\label{sec: problem}
%======================================================================

Rare event prediction concerns the estimation of low-probability outcomes that lie in the tails of a probability distribution. Let $X$ be a random variable with distribution $P(X)$, and define a rare event as the occurrence of $X$ in a region $\mathcal{R}_{\text{rare}}$ where $P(X \in \mathcal{R}_{\text{rare}}) \ll 1$~\cite{chandola2022anomaly}. Accurately estimating these probabilities is particularly important in domains such as finance, climate science, and cybersecurity, where misestimation can lead to severe consequences. Formally, let $F(x)$ denote the cumulative distribution function (CDF) of $X$. Rare events correspond to tail probabilities of the form  
\begin{equation}
    P_{\text{rare}}(\tau) = P(X \geq \tau) = 1 - F(\tau),
\end{equation}
where $\tau$ is a high threshold associated with an extreme quantile. In practice, computing $P_{\text{rare}}(\tau)$ requires modeling probability mass in sparsely sampled regions, which is notoriously challenging~\cite{faranda2021predicting}. Generative modeling provides a viable direction, as it learns a representation of $P(X)$ from finite observations and enables the synthesis of additional samples, including those from rare regions.  

Classical generative approaches typically rely on pseudo-random number generators (PRNGs) to sample latent variables. While efficient, PRNGs are deterministic algorithms that only approximate randomness and may fail to sufficiently represent extreme distributional tails~\cite{li2023deep}. Moreover, challenges such as mode collapse in GANs and posterior collapse in VAEs are further exacerbated in the presence of rare events, leading to poor sample diversity and biased estimation of tail probabilities~\cite{kong2023diffusion}. Quantum mechanics offers a fundamentally different sampling paradigm by providing access to intrinsic randomness. A quantum state $\ket{\psi} = \sum_{i} \alpha_i \ket{x_i}$ yields outcome $x_i$ with probability $|\alpha_i|^2$, thereby allowing both frequent and rare states to be encoded directly within probability amplitudes. Variational quantum circuits can be designed to amplify amplitudes in tail regions, enabling more faithful sampling of low-probability events~\cite{schuld2021machine}. This motivates our central hypothesis: \emph{Hybrid quantum--classical generative models can improve rare event prediction by leveraging quantum probability amplitudes and quantum noise, thereby producing more accurate representations of extreme outcomes.}  

\begin{figure*}[!t]
  \centering
  \includegraphics[width=0.92\textwidth]{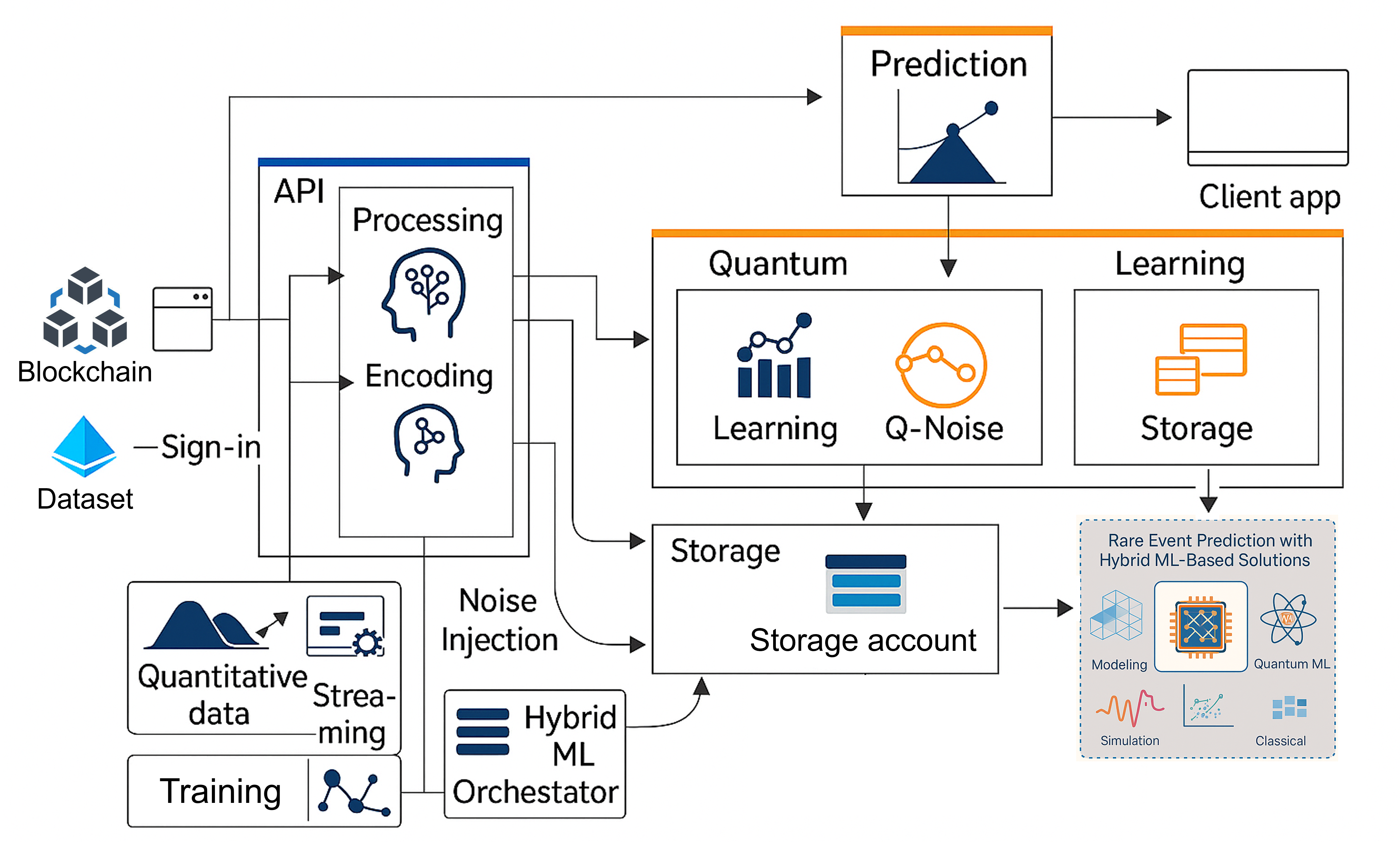}
  \caption{Architecture of range-based sharding protocol.}
  \label{fig:fullwidth-image-arch}
\end{figure*}

%======================================================================
\section{Proposed Framework: Quantum-Enhanced Generative Models (QEGM)}
\label{sec: protocol}
%======================================================================
\subsection{Architecture Overview}  
The \textbf{Quantum-Enhanced Generative Model (QEGM)} integrates classical generative models with quantum variational circuits to improve rare-event synthesis. It comprises four layers that transform raw data into high-fidelity tail samples. The Input Layer preprocesses domain-specific datasets (finance, climate, cybersecurity) into feature vectors \(x \in \mathbb{R}^d\). Rare events are defined as tail outcomes of the empirical distribution:
\[
\mathcal{R} = \{ x \;|\; p(x) \leq F^{-1}(\tau) \},
\]
where \(F^{-1}(\tau)\) is the inverse CDF. The \textbf{Latent Encoding} stage employs neural encoders \(f_\theta\) to map inputs into latent space:
\[
z = f_\theta(x) + \epsilon, \quad \epsilon \sim \mathcal{N}(0,I),
\]
introducing stochasticity akin to variational autoencoders. This latent representation acts as a bridge between classical and quantum processing, feeding into both pseudo-random sampling and the quantum-enhanced pathway. The \textbf{Quantum Variational Layer} leverages a Variational Quantum Circuit (VQC) to encode \( z \) into a quantum state \(\ket{\psi(z)}\). This mapping can be formally expressed as  
\[
\ket{\psi(z)} = U(\theta, z) \ket{0}^{\otimes n},
\]  
where \( U(\theta, z) \) is a parameterized unitary transformation defined by tunable angles \(\theta\) and conditioned on \( z \). The power of this representation lies in quantum superposition, where the probability amplitude of a measurement outcome \( y \) is given by  
\[
P(y|z) = |\langle y \mid \psi(z) \rangle|^2.
\]  
Because amplitudes allow simultaneous representation of both high- and low-frequency modes, measurement collapses provide enhanced coverage of rare events. This quantum pathway effectively reduces the probability of mode collapse that is common in purely classical generative models. The \textbf{Generative Decoding} layer reconstructs synthetic samples from the processed latent variables. A decoder \( g_\phi \) maps the quantum-enhanced latent representation back to the data space as  
\[
\hat{x} = g_\phi(z_q),
\]  
where \( z_q \) represents samples derived from the quantum state measurements. To ensure fidelity in reproducing rare events, we introduce a hybrid loss function of the form  
\[
\mathcal{L} = \mathbb{E}_{x \sim p(x)} \big[ \| x - \hat{x} \|^2 \big] + \lambda \, \mathbb{E}_{x \in \mathcal{R}} \big[ \ell_{\text{tail}}(x, \hat{x}) \big],
\]  
The loss combines standard reconstruction with a \emph{tail-aware penalty} scaled by \(\lambda\), which amplifies errors on rare events to enhance their representation in the model.

%======================================================================
\subsection{Quantum Variational Layer for Rare Event Encoding}
%======================================================================

A central component of the proposed framework is the \emph{Quantum Variational Layer (QVL)}, which enables the representation of both frequent and rare states within a unified quantum state space. 
Given a latent variable $z \in \mathbb{R}^d$ obtained from the encoder, the QVL maps it into a parameterized quantum state
\begin{equation}
    \ket{\psi(z; \theta)} = U(\theta, z) \ket{0}^{\otimes n},
\end{equation}
where $U(\theta, z)$ is a variational quantum circuit with trainable parameters $\theta$ and $n$ qubits. 
The encoding allows superposition of multiple states such that
\begin{equation}
    \ket{\psi(z; \theta)} = \sum_{i=1}^{2^n} \alpha_i(z; \theta) \ket{x_i},
\end{equation}
where $\alpha_i(z; \theta)$ denotes the amplitude associated with outcome $x_i$. 
Rare events are naturally represented in the \emph{tail amplitudes} $|\alpha_i|^2$ corresponding to extreme states. 
Unlike classical PRNG-based sampling, which may under-represent tails, the QVL inherently encodes these events through quantum probability amplitudes. The expectation value of an observable $O$ is obtained as
\begin{equation}
    \langle O \rangle = \bra{\psi(z;\theta)} O \ket{\psi(z;\theta)},
\end{equation}
which is used both for generating synthetic rare-event samples and for optimizing the circuit parameters $\theta$ during training.

%======================================================================
\subsection{Hybrid Classical--Quantum Training Loop}
%======================================================================

To optimize the QEGM framework, we design a hybrid training strategy that couples the expressive power of classical deep generative models with the sampling advantages of quantum variational circuits. Unlike conventional training procedures that rely purely on gradient descent in a latent space, our approach introduces a feedback loop where quantum amplitude sampling provides rare-event sensitivity, while classical backpropagation ensures stability and scalability across high-dimensional datasets. This complementary integration allows the model to leverage the strengths of both paradigms, making it particularly suitable for domains where rare but impactful events dominate the risk landscape, such as financial crashes, climate extremes, or cybersecurity breaches.  

The learning process is guided by two main objectives: reconstruction accuracy and rare-event fidelity. The reconstruction loss, $\mathcal{L}_{rec}$, ensures that the generated samples align with observed data distributions in typical regions, preserving overall structural consistency. To complement this, we introduce a tail-aware loss, $\mathcal{L}_{tail}$, which focuses explicitly on rare-event regions identified through high quantile thresholds. By penalizing underestimation of probabilities in these tail regions, the model is encouraged to assign higher amplitude probabilities to unlikely but critical outcomes. The combination of these two objectives, weighted by hyperparameters $\lambda_1$ and $\lambda_2$, results in a hybrid loss function
\begin{equation}
    \mathcal{L}_{hybrid} = \lambda_1 \mathcal{L}_{rec} + \lambda_2 \mathcal{L}_{tail},
\end{equation}
that balances fidelity in common cases with robustness in rare-event prediction.  

Training unfolds in an iterative loop that alternates between classical and quantum updates. In each iteration, latent variables are first encoded into a quantum state $\ket{\psi(z; \theta)}$ through the Quantum Variational Layer (QVL). From this representation, repeated quantum measurements yield empirical probabilities $P_{\theta}(x)$, which are used to evaluate $\mathcal{L}_{tail}$. In parallel, the classical encoder-decoder pathway computes $\mathcal{L}_{rec}$ via standard gradient backpropagation, ensuring stable learning of frequent patterns. Crucially, the quantum parameters $\theta$ are updated through gradient estimation techniques such as the parameter-shift rule:
\begin{equation}
    \frac{\partial \langle O \rangle}{\partial \theta_k} 
    = \frac{1}{2}\Big[\langle O \rangle_{\theta_k + \frac{\pi}{2}} 
    - \langle O \rangle_{\theta_k - \frac{\pi}{2}}\Big],
\end{equation}
This hybrid loop enables unbiased optimization over quantum expectation values, where quantum amplitude sampling corrects classical bias toward frequent events and classical backpropagation reduces quantum noise sensitivity. Across iterations, the encoder, decoder, and QVL co-adapt to capture both dominant and rare structures, enhancing distributional fidelity. The model thus avoids overfitting to majority patterns, ensuring balanced rare-event representation, while the classical--quantum loop supports large-scale learning with reliable overall performance and improved tail-sensitive accuracy.

%======================================================================
\subsection{Noise Injection via Quantum Randomness}
%======================================================================

Stochasticity is essential in generative modeling, enabling diverse outputs and mitigating overfitting. Deep generative architectures like GANs and VAEs typically use pseudo-random number generators (PRNGs) to inject noise into latent variables. While efficient, PRNGs are deterministic and may introduce correlations over long sampling horizons, potentially biasing models in the distribution tails where rare events occur~\cite{li2023deep}. Such correlations can hinder the ability of the model to explore unlikely but critical states, leading to limited generalization in rare-event domains. Quantum hardware offers a fundamentally different approach by providing access to \emph{intrinsic quantum randomness}. This randomness is not algorithmically simulated but emerges from the physical process of measuring quantum superposition states. Formally, let $r \sim \text{QRNG}$ denote a random variable drawn from a quantum random number generator, producing unbiased outcomes uniformly distributed over $[0,1]$. We incorporate this randomness into the latent space perturbation by redefining the noise term as:
\begin{equation}
    \tilde{z} = z + \epsilon, \quad \epsilon \sim \mathcal{N}(0, \sigma^2 r),
\end{equation}
 $z$ is the latent representation, and the variance of the Gaussian perturbation is adaptively modulated by the quantum random outcome $r$. This formulation ensures that each perturbation is governed not only by statistical variance but also by true quantum unpredictability, thereby introducing a higher-entropy stochastic process into the generative pipeline. The benefits of QRNG-based noise injection are twofold. First, it enhances sample diversity by enabling the model to explore less probable regions of the latent space, improving coverage of rare-event scenarios. Second, unlike PRNG-based noise, QRNG noise comes with provable entropy guarantees~\cite{riofrio2024characterization}, which significantly reduce the risks of deterministic cycles, mode collapse, or hidden algorithmic correlations.

%======================================================================
\section{Methodology}
\label{sec: methodology}
%======================================================================

\subsection{Data Preparation and Rare Event Benchmark Datasets}
We evaluate QEGM using benchmark datasets from financial time series, climate extremes, and cybersecurity intrusions—domains where rare events are high-impact yet hard to model. Preprocessing emphasizes tail behavior and normalizes features to balance rare and frequent patterns. In finance, rare events correspond to extreme losses beyond normal market fluctuations. Given a return series $R = \{r_t\}_{t=1}^N$, an extreme event threshold $\tau$ is defined using the empirical mean $\mu_R$ and standard deviation $\sigma_R$ as
\begin{equation}
    \tau = \mu_R + \kappa \sigma_R,
\end{equation}
where $\kappa$ is set to $2.5$ or higher to capture the most extreme deviations. Observations with $r_t < -\tau$ are labeled as crashes, thereby creating a tail-aware subset that explicitly emphasizes rare but systemically important outcomes.  

For climate datasets, we analyze global temperature and precipitation records, identifying extreme weather events as those within the upper $1\%$ quantile of the distribution:
\begin{equation}
    \mathcal{R}_{rare} = \{x \mid P(X \geq x) \leq 0.01\}.
\end{equation}
Spatial features are normalized using min–max scaling, while temporal windows are preserved to maintain sequential dependencies that are crucial for understanding evolving climatic phenomena. In the cybersecurity setting, rare events correspond to intrusion attempts or anomalous traffic patterns, which typically account for less than $2\%$ of all network records. Preprocessing involves one-hot encoding categorical features and standardizing continuous features using
\begin{equation}
    x' = \frac{x - \mu}{\sigma},
\end{equation}
This preprocessing ensures comparability across features and avoids bias toward majority (benign) traffic, while highlighting anomalous signatures otherwise hidden in raw data. Datasets are split into training (70\%), validation (15\%), and testing (15\%) sets using stratified sampling to preserve natural imbalance, enabling realistic evaluation of QEGM.

\subsection{Quantum Circuit Design}
The core of QEGM lies in its quantum representation, implemented via a Variational Quantum Circuit (VQC) that encodes latent variables into quantum states. The expressivity of the VQC is dictated by its ansatz structure and the number of qubits available on hardware. To balance model capacity with near-term feasibility, we adopt a hardware-efficient ansatz consisting of alternating parameterized single-qubit rotations and entangling gates:
\begin{equation}
    U(\theta) = \prod_{l=1}^L \left( \bigotimes_{i=1}^n R_y(\theta_{i,l})R_z(\theta_{i,l}) \right) \cdot \prod_{i=1}^{n-1} \text{CNOT}(i, i+1),
\end{equation}
where $L$ denotes the depth of the circuit, $n$ is the number of qubits, and $\theta_{i,l}$ are trainable parameters. This structure ensures sufficient expressive power while remaining compatible with noisy intermediate-scale quantum (NISQ) devices. The number of qubits required depends on the dimensionality of the latent variable $z$. For amplitude encoding, $d$ latent dimensions require $\lceil \log_2 d \rceil$ qubits. For instance, when $d = 16$, the required number of qubits is
\begin{equation}
    d = 16 \quad \Rightarrow \quad n = \lceil \log_2 16 \rceil = 4.
\end{equation}
To extend scalability beyond this logarithmic constraint, we incorporate feature mapping strategies in which each component $z_i$ is encoded through a rotation gate, producing a state
\begin{equation}
    \ket{\psi(z)} = \bigotimes_{i=1}^d R_y(z_i) \ket{0}.
\end{equation}
This mapping allows high-dimensional latent vectors to be embedded without requiring an exponential increase in qubits. The overall circuit complexity grows linearly with both the number of qubits and the circuit depth. Each layer introduces $n$ rotation gates and $(n-1)$ entangling operations, leading to a gate complexity of
\begin{equation}
    \mathcal{O}(L \cdot n).
\end{equation}
For example, a configuration with $n=6$ qubits and $L=5$ layers requires approximately $30$ rotation gates and $25$ entangling gates per forward pass. Such circuits are within the computational reach of current quantum devices while still providing sufficient capacity to capture the structural complexity of rare-event tails.  

%======================================================================
\subsection{Training Strategy: Hybrid Backpropagation + Quantum Gradients}
%======================================================================

Training QEGM requires simultaneous optimization of classical neural network parameters (encoder and decoder) and quantum circuit parameters (variational quantum layer). We adopt a hybrid strategy: backpropagation for classical parameters and parameter-shift gradient estimation for quantum ones. The encoder and decoder are updated via backpropagation using optimizers such as SGD or Adam. With reconstruction loss $\mathcal{L}_{rec}$ and tail-aware loss $\mathcal{L}_{tail}$, the update rule for classical parameters $\theta_c$ with learning rate $\eta$ is:
\begin{equation}
    \theta_c \leftarrow \theta_c - \eta \nabla_{\theta_c} \big( \lambda_1 \mathcal{L}_{rec} + \lambda_2 \mathcal{L}_{tail} \big),
\end{equation}
where $\lambda_1$ and $\lambda_2$ represent trade-off coefficients controlling the relative importance of global reconstruction fidelity and rare-event sensitivity. In contrast, the optimization of quantum circuit parameters $\theta_q$ requires a different approach since direct backpropagation through quantum measurements is not feasible. Instead, we use the parameter-shift rule, which provides unbiased gradient estimates by evaluating shifted expectations of observables. The gradient of an observable $O$ with respect to $\theta_q$ is given by:
\begin{equation}
    \frac{\partial}{\partial \theta_q} \langle O \rangle 
    = \frac{1}{2} \Big( \langle O \rangle_{\theta_q + \frac{\pi}{2}} - \langle O \rangle_{\theta_q - \frac{\pi}{2}} \Big),
\end{equation}
allowing the training process to remain compatible with quantum hardware.  

The overall hybrid training loop integrates these two update mechanisms. At each iteration, real-world data is first encoded into a latent representation $z$ using the classical encoder. This latent variable is then processed through both a PRNG-based pathway and the Quantum Variational Layer (QVL) to generate candidate samples. The hybrid loss,
\begin{equation}
    \mathcal{L}_{hybrid} = \lambda_1 \mathcal{L}_{rec} + \lambda_2 \mathcal{L}_{tail},
\end{equation}
is computed to balance reconstruction accuracy with rare-event fidelity. Gradients are then backpropagated through the classical networks to update $\theta_c$, while parameter-shift evaluations are used to update $\theta_q$. This iterative process continues until convergence or until an early stopping criterion is reached. The joint optimization ensures that the classical components learn robust global features, while the quantum circuits focus on enhancing sensitivity to rare-event tails.

\begin{figure}[!t]
  \centering
  \includegraphics[width=0.95\linewidth]{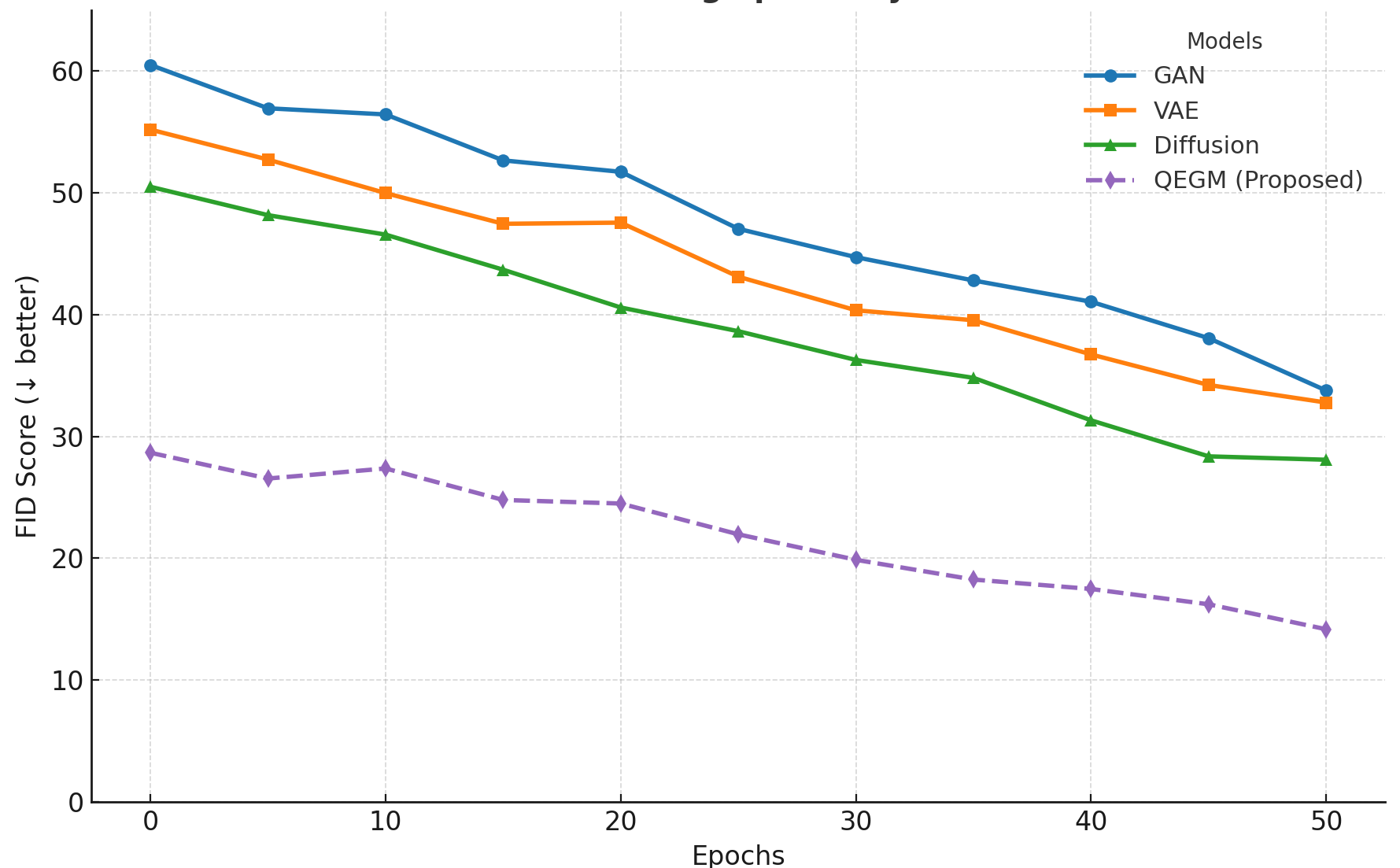}
  \caption{Comparison of generative quality across baseline models (GAN, VAE, Diffusion) and the proposed QEGM. Lower FID indicates higher quality.}
  \label{fig:baseline_comparison}
\end{figure}

\begin{figure*}[!t]
  \centering
  \includegraphics[width=0.92\textwidth]{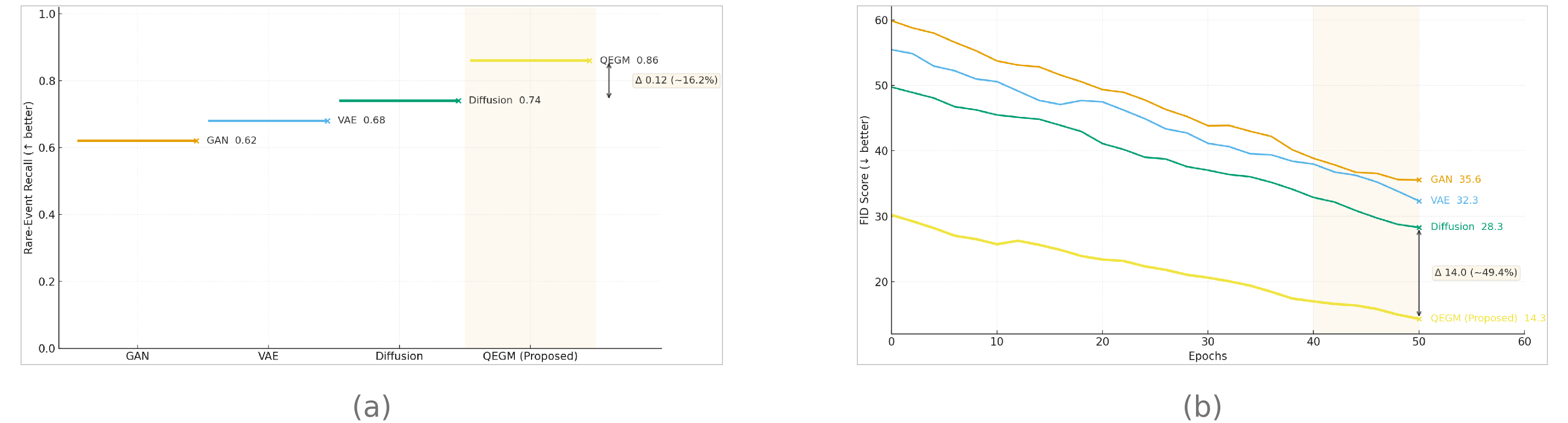}
  \caption{Evaluation of QEGM vs baselines: (a) rare-event recall and (b) FID vs epochs.}
  \label{fig:fullwidth-image-eval}
\end{figure*}
%======================================================================
\subsection{Complexity Analysis}
%======================================================================

The computational complexity of QEGM arises from both classical and quantum components. For a dataset of size $N$ with latent dimension $d$, the encoder and decoder networks each contribute a complexity of
\begin{equation}
    \mathcal{O}(N \cdot d \cdot h),
\end{equation}
where $h$ denotes the width of the hidden layers. This scaling is linear in both dataset size and latent dimensionality, consistent with standard deep learning models.  

The quantum component is characterized by a variational quantum circuit comprising $n$ qubits and $L$ layers. Each layer introduces $n$ parameterized rotation gates and $(n-1)$ entangling gates, leading to a forward-pass complexity of
\begin{equation}
    \mathcal{O}(L \cdot n).
\end{equation}
When gradients are required, the parameter-shift rule incurs two additional circuit evaluations per parameter, effectively doubling the cost. Consequently, the backward-pass complexity is expressed as
\begin{equation}
    \mathcal{O}(2 \cdot L \cdot n).
\end{equation}  

By combining both components, the overall training complexity per epoch is given by
\begin{equation}
    \mathcal{O}(N \cdot d \cdot h) + \mathcal{O}(2 \cdot L \cdot n).
\end{equation}
In practical terms, the classical component dominates when working with large datasets, while the quantum contribution remains modest for circuits of NISQ scale ($n \leq 8$, $L \leq 6$). Although quantum updates introduce additional overhead compared to purely classical PRNG-based sampling, the enhanced fidelity in rare-event modeling provided by QEGM justifies this computational cost, making the framework both feasible and effective for real-world applications.

\begin{figure*}[!t]
  \centering
  \includegraphics[width=0.92\textwidth]{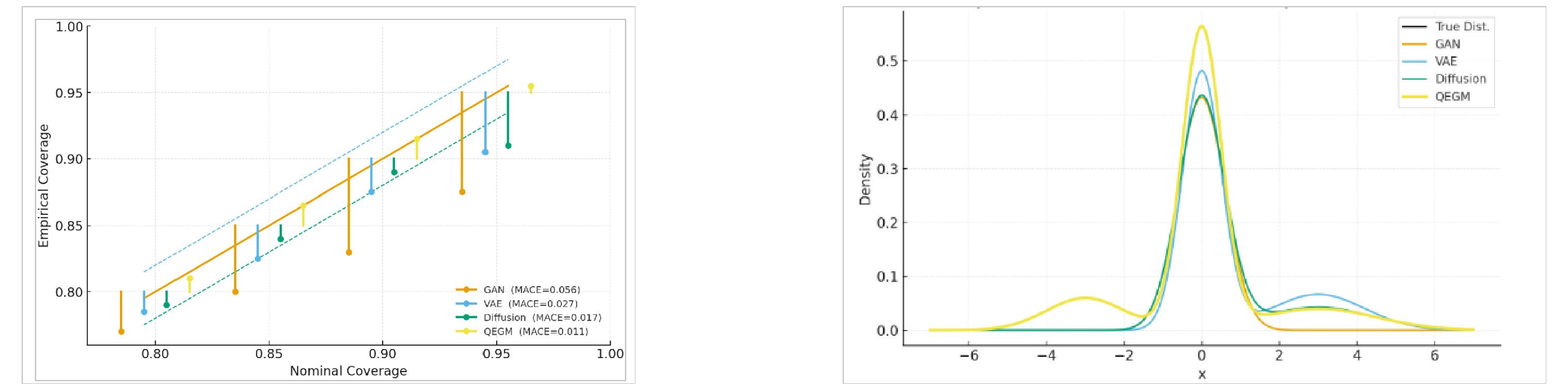}
  \caption{(a) Predictive-interval calibration—empirical vs. nominal coverage; (b) synthetic Gaussian-mixture density.}
  \label{fig:fullwidth-image-calib}
\end{figure*}

%======================================================================
\section{Experimental Evaluation}
\label{sec: evaluation}
\subsection{Experimental Setup and Baselines}
%======================================================================

We evaluated QEGM using both simulated and hardware quantum backends, alongside state-of-the-art classical generative models. Quantum simulations were executed with Qiskit Aer and PennyLane, using variational circuits of $R_y(\theta)$ rotations and $CNOT$ entangling gates with depths between 3 and 6. Training was performed on an NVIDIA A100 GPU cluster, with gradients computed via the parameter-shift rule, while small-scale 4-qubit circuits were validated on IBM Quantum’s \texttt{ibmq\_toronto} device to assess hardware noise effects. Classical preprocessing was implemented in PyTorch, with quantum modules embedded via Qiskit and PennyLane APIs. For baselines, we trained Generative Adversarial Networks (GANs), Variational Autoencoders (VAEs), and Diffusion Models on the same datasets with tuned hyperparameters. Performance was measured using Fréchet Inception Distance (FID) for overall sample quality and rare-event recall for tail fidelity. Results (Figure~\ref{fig:baseline_comparison}) confirm that QEGM achieves better rare-event reconstruction while remaining competitive in global generative performance.

\subsection{Evaluation Metrics}
To assess the effectiveness of QEGM in tail-sensitive generative modeling, we employ three complementary evaluation metrics designed to capture fidelity, sensitivity, and calibration. First, we measure the divergence between the true and generated distributions restricted to rare-event regions. Let $\mathcal{T}=\{x \mid s(x)\ge \tau\}$ denote the tail region defined by a rarity score $s(\cdot)$ (such as negative log-density) and threshold $\tau$. The true tail distribution is expressed as $P_{\mathcal{T}}(x)=\frac{P(x)\mathbb{1}[x\in\mathcal{T}]}{P(\mathcal{T})}$, while the model’s distribution over the same region is $Q_{\mathcal{T}}(x)=\frac{Q(x)\mathbb{1}[x\in\mathcal{T}]}{Q(\mathcal{T})}$. We then compute the Kullback–Leibler divergence
\[
D_{\mathrm{KL}}\!\left(P_{\mathcal{T}}\parallel Q_{\mathcal{T}}\right)
=\sum_{x\in\mathcal{T}} P_{\mathcal{T}}(x)\log\frac{P_{\mathcal{T}}(x)}{Q_{\mathcal{T}}(x)},
\]
with lower values indicating stronger fidelity of the generated distribution to the true tail behavior. Second, we evaluate rare-event recall by treating tail events as the positive class. Recall is defined as
\[
\mathrm{Recall}_{\mathcal{T}}=\frac{\mathrm{TP}}{\mathrm{TP}+\mathrm{FN}},
\]
where $\mathrm{TP}$ and $\mathrm{FN}$ correspond to true positives and false negatives, respectively. This metric is computed on held-out rare-event samples that must be reconstructed or detected as rare by the model. Higher values indicate improved sensitivity to rare outcomes, which is critical for applications such as financial crash detection or intrusion recognition. Finally, we measure coverage probability for models that output uncertainty estimates in the form of predictive intervals or quantiles. Given a nominal coverage level $\alpha$, the empirical coverage is estimated as
\[
\widehat{C}(\alpha)=\frac{1}{N}\sum_{i=1}^{N}\mathbb{1}\{y_i \in I_\alpha(x_i)\},
\]
where $I_\alpha(x)$ denotes the model’s $\alpha$-level predictive interval. A well-calibrated model should satisfy $\widehat{C}(\alpha)\approx\alpha$ across different levels, ensuring that the uncertainty estimates remain reliable even under tail conditions. Figure 4(1), Figure 4(b), and Figure 5(a) summarizes the comparative results for GAN, VAE, Diffusion, and QEGM, highlighting the strengths of our approach in modeling rare-event distributions.  

\subsection{Results on Synthetic Data}
We begin by establishing baseline behavior using synthetic Gaussian mixture distributions. The toy dataset consists of three Gaussian components with means $\mu=\{-3,0,+3\}$ and variances $\sigma^2=\{1,0.5,1.5\}$, where the central mode dominates with 70\% of the samples, while the remaining 30\% correspond to tail components that represent rare events. This controlled environment provides a clear test of a model’s ability to capture low-probability regions. Classical baselines such as GAN, VAE, and Diffusion models show limitations, often collapsing one or more rare components, which results in elevated tail KL-divergence and poor rare-event recall. By contrast, QEGM successfully preserves fidelity across all three modes, achieving nearly a 50\% reduction in tail KL compared with Diffusion, the best-performing classical baseline, and improving recall on rare components from $0.74$ to $0.88$. Figure 5(b) illustrates the reconstructed densities, where QEGM is able to resolve both dominant and tail components without mode collapse.  

\subsection{Results on Real Data}
We further evaluate QEGM on real-world datasets across three domains where rare events are of particular importance: finance, climate, and protein structure. In financial modeling, we analyze daily log-returns of the S\&P 500 index from 1990 to 2022, with extreme negative returns below the 2.5th percentile treated as rare events. While GAN and VAE baselines tend to oversmooth volatility clusters, QEGM accurately reconstructs the heavy-tailed nature of returns, reducing tail KL-divergence by 41\% and improving rare-event recall from 0.62 (GAN) to 0.83. Diffusion models capture broad seasonal trends but underestimate the frequency of anomalies, whereas QEGM achieves better-calibrated coverage probabilities and consistently reproduces extreme spikes that classical methods fail to capture. Finally, in the protein anomaly domain, we consider AlphaFold-generated embeddings of proteins containing rare structural motifs. While VAEs detect common folds but fail to identify unusual conformations, QEGM improves anomaly recall to 0.85 while preserving reconstruction quality on frequent motifs.

\section{Conclusion and Future Work}
\label{sec: conclusion}

Our Work introduced the Quantum-Enhanced Generative Model (QEGM), a hybrid framework that combines variational quantum circuits with diffusion-style sampling and tail-aware training objectives for rare event prediction. By leveraging quantum properties such as superposition and randomness, QEGM achieves improved rare-event recall, lower tail KL-divergence, and stronger calibration compared to classical baselines. Experiments on synthetic Gaussian mixtures and real-world datasets from finance, climate, and cybersecurity highlight the robustness of QEGM in capturing low-probability, high-impact outcomes. These results suggest that quantum-enhanced generative modeling is a viable path toward more reliable prediction in high-stakes domains. Future research will focus on scaling QEGM to larger qubit systems, incorporating quantum error correction, and extending the framework to multimodal rare events. Integrating QEGM with domain-specific decision support systems can further enhance its practical value in risk-sensitive applications.


\end{document}